\documentclass{article}
\usepackage{iclr2018_conference,times}
\usepackage{hyperref}
\usepackage{url}

\usepackage{times}
\usepackage{graphicx} 
\usepackage{subfigure} 

\usepackage{booktabs}

\usepackage{natbib}

\usepackage[utf8]{inputenc}
\usepackage[T1,T5]{fontenc}

\usepackage{algorithm}
\usepackage{algorithmic}
\usepackage{multirow,multicol}
\usepackage{amsmath}
\usepackage{amssymb}
\usepackage{mdwlist}

\usepackage{tikz}
\usetikzlibrary{shapes,arrows}

\newcommand{\setS}{\mathcal{S}}

\iclrfinalcopy 

\title{Towards Neural Phrase-based Machine \\ Translation}

\author{Po-Sen Huang$^\star$, Chong Wang\thanks{Work performed while CW, DZ, and LD were at Microsoft Research and SH was interning at Microsoft Research.}$^{\hspace{0.015in}~\dagger}$, Sitao Huang\footnotemark[1]$^{\hspace{0.015in}~\ddagger}$, Dengyong Zhou\footnotemark[1]$^{\hspace{0.015in}~\dagger}$, Li Deng\footnotemark[1]$^{\hspace{0.015in}~\diamond}$ \\
	$^\star$Microsoft Research, $^\dagger$Google, $^\ddagger$University of Illinois at Urbana-Champaign, $^\diamond$Citadel\\
	{\small\texttt{pshuang@microsoft.com, \{chongw, dennyzhou\}@google.com},}\\ {\small\texttt{shuang91@illinois.edu, l.deng@ieee.org}}
	\\
}

\begin{document} 
\maketitle

\begin{abstract} 
  In this paper, we present Neural Phrase-based Machine Translation (NPMT).\footnote{The source code is available at \url{https://github.com/posenhuang/NPMT}.} Our
  method explicitly models the phrase structures in output sequences using
  Sleep-WAke Networks (SWAN), a recently proposed segmentation-based sequence
  modeling method. To mitigate the monotonic alignment requirement of SWAN, we
  introduce a new layer to perform (soft) local reordering of input sequences.
  Different from existing neural machine translation (NMT) approaches, NPMT does
  not use attention-based decoding mechanisms. 
  Instead, it directly outputs phrases in a sequential order and can decode in linear time. 
  Our experiments show that NPMT achieves superior performances on IWSLT 2014 German-English/English-German and IWSLT
  2015 English-Vietnamese machine translation tasks compared with strong NMT baselines.
  We also observe that our method produces meaningful phrases in output
  languages.
\end{abstract} 

\section{Introduction}
\label{sec:intro}

A word can be considered as a basic unit in languages. However, in many cases, we
often need a phrase to express a concrete meaning.
For example, consider understanding the following sentence, 
``machine learning is a field of computer science''. It may become easier to comprehend if we segment it
as ``[machine learning] [is] [a field of] [computer science]'', where the
words in the bracket `[]' are regarded as ``phrases''.
These phrases have their own meanings,  and can often be reused in other contexts. 

The goal of this paper is to explore the use of phrase structures aforementioned
for neural network-based machine translation
systems~\citep{Sutskever:2014,Bahdanau:2014}.
To this end, we develop a neural machine translation method that explicitly
models phrases in target language sequences. Traditional
phrase-based statistical machine translation (SMT) approaches have been shown to
consistently outperform word-based
ones~\citep{koehn2003statistical,koehn2009statistical,lopez2008statistical}. However, modern neural machine translation (NMT)
methods~\citep{Sutskever:2014,Bahdanau:2014,luong2015effective} do not have an
explicit treatment on phrases, but they still work surprisingly well and have
been deployed to industrial systems~\citep{TACL863,wu2016google}. 
The proposed Neural Phrase-based Machine Translation (NPMT) method tries to
explore the advantages from both kingdoms. It builds upon Sleep-WAke Networks
(SWAN), a segmentation-based sequence modeling technique described
in~\cite{wang2017sequence}, where segments (or phrases) are automatically
discovered given the data. However, SWAN requires monotonic alignments between
inputs and outputs. This is often not an appropriate assumption in many language
pairs. To mitigate this issue, we introduce a new layer to perform (soft) local
reordering on input sequences. Experimental results show that NPMT outperforms
attention-based NMT baselines in terms of the BLEU
score~\citep{papineni2002bleu} on IWSLT 2014 German-English/English-German and
IWSLT 2015 English-Vietnamese translation tasks. We believe our method is one
step towards the full integration of the advantages from neural machine
translation and phrase-based SMT.

This paper is organized as follows. Section~\ref{sec:model} presents the neural
phrase-based machine translation model. Section~\ref{sec:experiment}
demonstrates the usefulness of our approach on several language pairs. We
conclude our work with some discussions in Section~\ref{sec:conclusion}.

\tikzstyle{block} = [rectangle,draw,text width=10em, text centered, rounded corners, minimum height=2em]
\tikzstyle{textblock} = [rectangle, text width=10em, text centered, rounded corners, minimum height=1em]
\tikzstyle{line} = [draw, -latex']

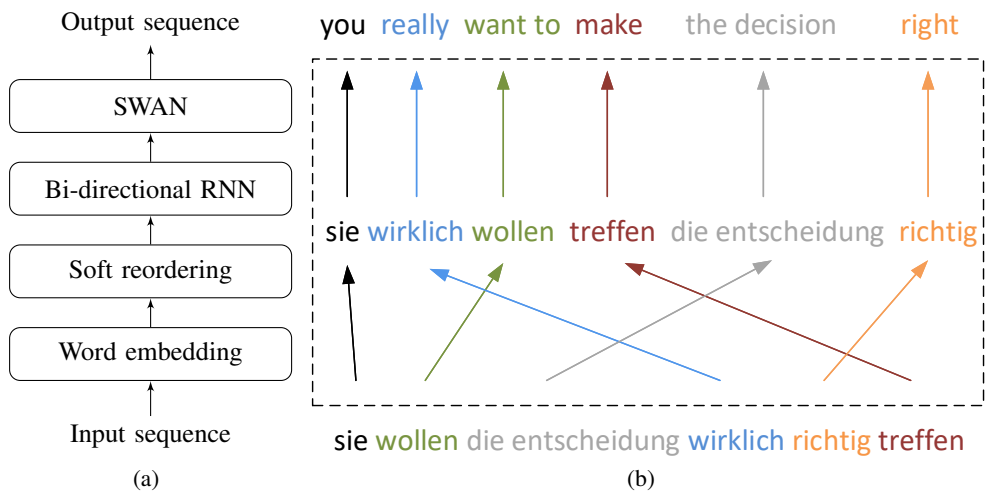
\begin{figure} \centering
	\subfigure[]{\label{fig:npmt_architecture}\begin{tikzpicture}[node distance = 1.1cm, auto]
		\node [textblock] (output) {Output sequence}; 
		\node [block, below of = output] (swan){SWAN};
		\node [block, below of=swan] (rnn) {Bi-directional RNN};
		\node [block, below of=rnn] (reordering) {Soft reordering};
		\node [block, below of= reordering] (embedding) {Word embedding};
		\node [textblock, below of = embedding] (input){Input sequence}; 
		
		\path [line] (swan) -- (output); 
		\path [line] (rnn) -- (swan);
		\path [line] (reordering) -- (rnn) ;
		\path [line] (embedding) -- (reordering) ;
		\path [line] (input) -- (embedding); 
		\end{tikzpicture}}
  \subfigure[]{\label{fig:npmt_example}\includegraphics[width=0.66\columnwidth]{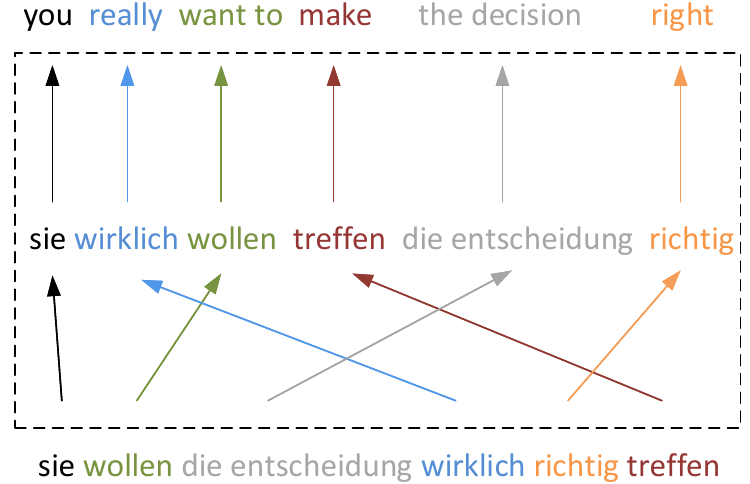}}
  \caption{\small{(a) The overall architecture of NPMT. (b) An illustration of using
  NPMT in German-English translation. Ideally, phrases in the source sentence
  (German) are first reordered. Given the new order, phrases can be translated
  one by one to the target phrases. These translated phrases then compose the
  target sentence (English). Phrase boundaries in the target language are not
  predefined, but automatically discovered by the model. No attention-based
  decoders are used here.}}
	\label{fig:npmt}
\end{figure}

%
%

\section{Neural phrase-based machine translation}
\label{sec:model}

We first give an overview of the proposed NPMT architecture and some related
work on incorporating phrases into NMT. We then describe the two key building
blocks in NPMT: 1) SWAN, and 2) the soft reordering layer which alleviates the
monotonic alignment requirement of SWAN. In the context of machine translation,
we use ``segment'' and ``phrase'' interchangeably.

\subsection{The overall architecture of NPMT}
Figure~\ref{fig:npmt_architecture} shows the overall architecture of NPMT. The
input sequence is first turned into embedding representations and then they go
through a (soft) reordering layer (described below in
Section~\ref{sec:reorder}). We then pass these ``reordered'' activations to the
bi-directional RNN layers, which are finally fed into the SWAN layer to directly
output target language in terms of segments (or phrases). While it is possible
to replace bi-directional RNN layers with other
layers~\citep{gehring2017convolutional}, in this paper, we have only explored
this particular setting to demonstrate our proposed idea.

There have been several works that propose different ways to incorporate phrases
into attention-based neural machine translation, such
as~\cite{Tang2016neural,Wang:2017emnlp,dahlmann-EtAl:2017:EMNLP2017}. These
approaches typically use predefined phrases (obtained by external methods, e.g.,
phrase-based SMT) to guide or modify the existing attention-based decoder. The
major difference from our approach is that, in NPMT, we do not use 
attention-based decoding mechanisms, and our phrase structures for the target
language are automatically discovered from the training data. Another line of
related work is the segment-to-segment neural transduction model (SSNT)
~\citep{yu-buys-blunsom:2016:EMNLP2016}, which shows promising results on a
Chinese-to-English translation task under a noisy channel
framework~\citep{yu2016neural}. In SSNT, the segments are implicit, and the
monotonic alignments between the inputs and outputs are achieved using latent
variables. The latent variables are marginalized out during training using dynamic programming.

\subsection{Modeling phrases with SWAN}
\label{sec:swan}
Here we review the SWAN model proposed in~\cite{wang2017sequence}. SWAN defines
a probability distribution for the output sequence given an input sequence. It
models all valid output segmentations of the output sequence as well as the
monotonic alignments between the output segments and the input sequence. Empty
segments are allowed in the output segmentations. It does not make any
assumption on the lengths of input or output sequence. 

\begin{figure}[t]
\begin{center}
\vskip -0.2in
\centerline{\includegraphics[width=0.48\columnwidth]{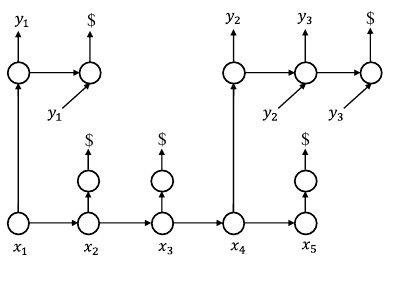}}
\vskip -0.2in
\caption{\small{Courtesy to~\citet{wang2017sequence}.  Symbol $\$$ indicates the end of
  a segment. Given a sequence of inputs $x_1, \ldots, x_5$, which is from the outputs from the bi-directional RNN of Figure \ref{fig:npmt_architecture}, SWAN emits one particular segmentation of $y_{1:3}=\pi(a_{1:5})$,
  where $\{a_1=\{y_1, \$\}, a_2=\{\$\}, a_3=\{\$\}, a_4=\{y_2, y_3, \$\},
  a_5=\{\$\}\}$.  Here $x_1$ wakes (emitting segment $a_1$) and $x_4$ wakes
  (emitting segment $a_4$) while $x_2$, $x_3$ and $x_5$ sleep (emitting empty
  segments $a_2$, $a_3$ and $a_5$ respectively).}}
\label{fig:wasm}
\end{center}
\end{figure}

Assume input sequence for SWAN is $x_{1:T'}$, which is the outputs from bi-directional RNN of Figure \ref{fig:npmt_architecture}, 
and output sequence is $y_{1:T}$. Let
$\setS_y$ denote the set containing all valid segmentations of $y_{1:T}$, with
the constraint that the number of segments in a segmentation is the same as the
input sequence length, $T'$. Let $a_t$ denote a segment or phrase in the
target sequence. Empty segments are allowed to ensure that we can correctly
align segment $a_t$ to input element $x_t$. Otherwise, we might not have
a valid alignment for the input and output pair. See Figure~\ref{fig:wasm} for
an example of the emitted segmentation of $y_{1:T}$.
The probability of the sequence $y_{1:T}$ is  defined as the sum of the
probabilities of all the segmentations in $\setS_y \triangleq   \{a_{1:T'}:
\pi(a_{1:T'}) = y_{1:T}\}$,\footnote{If predefined phrase structure information
is provided for the target language in advance, we can incorporate it into SWAN
by restricting the size of $\setS_y$. We leave the exploration of this option as
future work.}
\begin{align}
 p(y_{1:T} |x_{1:T'}) \triangleq \sum_{a_{1:T'} \in \setS_y}
  \prod_{t=1}^{T'} p(a_t|x_t), \label{eq:model-2}
\end{align} 
where the $p(a_t|x_t)$ is the segment probability given input element $x_t$,
which is modeled using a recurrent neural network (RNN) with an additional
softmax layer. $\pi(\cdot)$ is the concatenation operator and the symbol $\$$,
end of a segment, is ignored in the concatenation operator $\pi(\cdot)$.  (An
empty segment, which only contains $\$$ will thus be ignored as well.) SWAN can
be also understood via a generative model,
\begin{enumerate}
  \item For $t=1,...,T'$:
    \begin{enumerate}
      \item Given an initial state of $x_t$, sample words from RNN until we reach
        an end of segment symbol $\$$. This gives us a segment $a_t$.
    \end{enumerate}
  \item Concatenate $\{a_1,..., a_{T'}\}$ to obtain the output sequence via
  $\pi(a_{1:T'}) = y_{1:T}$.
\end{enumerate}
Since there are more than one way to obtain the same $y_{1:T}$ using the
generative process above, the probability of observing $y_{1:T}$ is obtained by summing
over all possible ways, which is Eq.~\ref{eq:model-2}.

Note that $|\setS_y|$ is exponentially large, direct summation quickly becomes
infeasible when $T$ or $T'$ is not small. Instead, \citet{wang2017sequence}
developed an exact
dynamic programming algorithm to tackle the computation challenges.\footnote{The
computational complexity of SWAN is still high even with the dynamic programming
algorithm.  This is the reason that it takes a longer time to train our
method for larger datasets such as in WMT translation tasks (weeks for a
moderate model size).  In the meantime, we are actively looking into the
algorithms that can significantly speed up SWAN.} The key idea is that although
the number of possible segmentations is exponentially large, the number of
possible segments is polynomial---$O(T^2)$. In other words, it is possible to
first compute all possible segment probabilities, $p(a_t |x_t)$, $\forall a_t,
x_t$, and then use dynamic programming to calculate the output sequence
probability $p(y_{1:T} |x_{1:T'})$ in Eq.~\eqref{eq:model-2}. The feasibility of
using dynamic programming is due to a property of segmentations---a segmentation
of a subsequence is also part of the segmentation of the entire sequence. In
practice, a maximum length $L$ for a segment $a_t$ is enforced to reduce the
computational complexity, since the length of useful segments is often not very
long. 
\cite{wang2017sequence} also discussed a way to carry over information
across segments using a separate RNN, which we will not elaborate here. 
We refer the readers to the original paper for the algorithmic details.

SWAN defines a conditional probability for an output sequence given an input
one. It can be used in many sequence-to-sequence tasks. In practice, a sequence
encoder like a bi-directional RNN can be used to process the raw input sequence
(like speech signals or source language) to obtain $x_{1:T'}$ that is to be
passed into SWAN for decoding. For example, \cite{wang2017sequence} demonstrated
the usefulness of SWAN in the context of speech recognition.

Greedy decoding for SWAN is straightforward. We first note that $p(a_t|x_t)$ is
modeled as an RNN with an additional softmax layer. 
Given each $p(a_t|x_t)$, $\forall t\in {1, \ldots, T'}$, is independent of each other, we can run the RNN in parallel to produce an output segment (possibly empty) for each $p(a_t|x_t)$. 
We then concatenate these output segments to form the greedy
decoding of the entire output sequence.  
The decoding satisfies the non-autoregressive property \citep{gu2018non-autoregressive} and the decoding complexity is $O(T'L)$.
See~\cite{wang2017sequence} for the
algorithmic details of the beam search decoder.

We finally note that, in SWAN (thus in NPMT), only output segments are explicit;
input segments are implicitly modeled by allowing empty segments in the
output.  This is conceptually different from the traditional phrase-based SMT
where both inputs and outputs are phrases (or segments). We leave the option of
exploring explicit input segments as future work.

\subsection{Local reordering of input sequences}
\label{sec:reorder}

SWAN assumes a monotonic alignment between the output segments and the input
elements. For speech recognition experiments in~\citet{wang2017sequence}, this
is a reasonable assumption. However, for machine translation, this is usually too
restrictive.  In neural machine translation literature, attention mechanisms were
proposed to address alignment
problems~\citep{Bahdanau:2014,luong2015effective,raffel2017online,vaswani2017attention}.
But it is not clear how to apply a similar attention mechanism to SWAN due to
the use of segmentations for output sequences.

One may note that in NPMT, a bi-directional RNN encoder for the
source language can partially mitigate the alignment issue for SWAN, since it can access
every source word. However, from our empirical studies, it is not enough to obtain the best
performance.  
Here we augment SWAN with a reordering layer that does (soft) local 
reordering of the input sequence.  
This new model leads to promising
results on the IWSLT 2014 German-English/English-German, and IWSLT 2015
English-Vietnamese machine translation tasks. One additional advantage of using
SWAN is that since SWAN does not use attention mechanisms, decoding can be done in parallel with linear complexity, as now we remove the need to query the entire input source for every output
word~\citep{raffel2017online, gu2018non-autoregressive}.

\begin{figure} \centering
	\subfigure[]{\label{fig:reorder_a}\includegraphics[width=0.40\columnwidth]{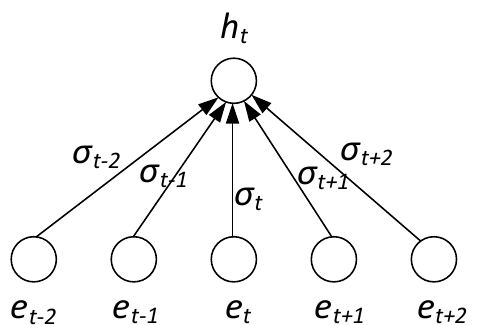}}
	\subfigure[]{\label{fig:reorder_b}\includegraphics[width=0.29\columnwidth]{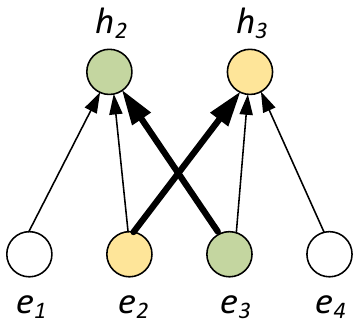}}
  \caption{\small{(a) Example of a local reordering layer of window size 5 ($\tau=2$) to
  compute $h_t$. Here $\sigma_{t-2+i}\triangleq
  \sigma(w_i^T[e_{t-2};e_{t-1};e_t; e_{t+1}; e_{t+2}])$, $i=0,\dots,4$, are the
  gates that decides how much information $h_t$ should accept from those
  elements from this input window. Note that all information available in this
  input window helps decides each gate. (b) An illustration of the reordering
  layer that swaps information between $e_2$ and $e_3$ and contributes to $h_3$
  and $h_2$, respectively.}}
	\label{fig:reordering}
\end{figure}

We now describe the details of the local reordering layer shown in
Figure~\ref{fig:reorder_a}. Denote the input to the local reordering layer by
$e_{1:T'}$, which is the output from the word embedding layer of Figure \ref{fig:npmt_architecture}, 
and the output of this layer by $h_{1:T'}$, which is fed as inputs to the bi-directional RNN of Figure \ref{fig:npmt_architecture}.
We compute $h_t$ as
\begin{equation}
\label{eq:reordering}
 h_t = {\rm tanh}\left(\sum_{i=0}^{2\tau} \sigma\left(w_i^T
[e_{t-\tau};\ldots;e_t; \ldots; e_{t+\tau} ]\right) e_{t-\tau+i} \right).
\end{equation}
where $\sigma(\cdot)$ is the sigmoid function, and $2\tau+1$ is the local
reordering window size. Notation $[e_{t-\tau};\ldots;e_t; \ldots; e_{t+\tau}]$
is the concatenation of vectors $e_{t-\tau}, \ldots,e_t, \ldots, e_{t+\tau}$.
For $i=0,\ldots, 2\tau$, notation $w_i$ is the parameter for the gate function at
position $i$ of the input window. It decides the weight of $e_{t-\tau+i}$
through the gate $\sigma\left(w_i^T [e_{t-\tau};\ldots;e_t; \ldots;
e_{t+\tau}]\right)$. The final output $h_t$ is a weighted linear combination of
the input elements, $e_{t-\tau}, \ldots, e_t, \ldots, e_{t+\tau}$, in the window
followed by a nonlinear transformation by the ${\rm tanh}(\cdot)$ function.


Figure~\ref{fig:reorder_b} illustrates how local reordering works. Here we want
to (softly) select an input element from a window given all information available
in this window.  Suppose we have two adjacent windows, $(e_1, e_2,e_3)$ and
$(e_2, e_3, e_4)$. 
If $e_3$ gets the largest weight ($e_3$ is {\it picked}) in the first window and $e_2$ gets the largest weight ($e_2$ is {\it picked}) in the second window,
$e_2$ and $e_3$ are effectively reordered.
Our layer is different from
the attention
mechanism~\citep{Bahdanau:2014,luong2015effective,raffel2017online,vaswani2017attention}
in following ways.  
First, we do not have a query to begin with as in standard attention mechanisms. 
Second, unlike standard attention, which is top-down from a decoder state to encoder states, the reordering operation is bottom-up. 
Third, the weights $\{w_i\}_{i=0}^{2\tau}$ capture the relative positions of the input elements, whereas the weights are the same for different queries and encoder hidden states in the attention mechanism (no positional information). 
The reordering layer performs locally similar to a convolutional layer and the positional information is encoded by a different parameter $w_i$ for each relative position $i$ in the window. 
Fourth, we do not normalize the weights for the input elements $e_{t-\tau},\ldots, e_t, \ldots, e_{t+\tau} $. 
This provides the reordering capability and can potentially turn off everything if needed. 
Finally, the gate of any position $i$ in the reordering window is determined by all input elements $e_{t-\tau},\ldots, e_t, \ldots, e_{t+\tau}$ in the window. 
We provide a visualizing example of the reordering layer
gates that performs input swapping in Appendix \ref{sec:reordering_analysis}.

One related work to our proposed reordering layer is the Gated Linear Units
(GLU)~\citep{dauphin2016language} which can control the information flow of the
output of a traditional convolutional layer. But GLU does not have a mechanism
to decide which input element from the convolutional window to choose. From our
experiments, neither GLU nor traditional convolutional layer helped our NPMT.
Another related work to the window size of the reordering layer is the distortion limit in traditional phrase-based statistical machine translation methods \citep{brown1993mathematics}. 
Different window sizes restrict the context of each position to different numbers of neighbors. 
We provide an empirical comparison of different window sizes in Appendix \ref{sec:window_reorder}. 


\section{Experiments}
\label{sec:experiment}

In this section, we evaluate our model on the IWSLT 2014
German-English~\citep{cettolo2014report}, IWSLT 2014 English-German, and IWSLT 2015
English-Vietnamese~\citep{cettolo2015iwslt} machine translation tasks. We note that, in this paper, we
limit the applications of our model to relatively small datasets to demonstrate
the usefulness of our method. We plan to conduct more large scale experiments in
future work.

\subsection{IWSLT14 German-English}
\label{sec:iwslt_de-en}
We evaluate our model on the German-English machine translation track of the IWSLT 2014
evaluation campaign~\citep{cettolo2014report}.
The data comes from translated TED talks, and the dataset contains
roughly 153K training sentences, 7K development sentences, and 7K test
sentences. We use the same preprocessing and dataset splits as in
\citet{DBLP:journals/corr/RanzatoCAZ15,
wiseman2016sequence,Bahdanau2016}. 
The German and English vocabulary sizes are 32,010 and 22,823 respectively.

We report our IWSLT 2014 German-English experiments using one reordering layer
with window size 7, two layers of bi-directional GRU encoder (Gated recurrent unit,
\cite{Chung:2014}) with 256 hidden units, and two layers of unidirectional GRU decoder
with 512 hidden units.  We add dropout with a rate of $0.5$ in the GRU layer.
We choose GRU since baselines for comparisons were using GRU.  The maximum
segment length is set to 6.  Batch size is set as 32 (per GPU) and the Adam
algorithm~\citep{kingma2014adam} is used for optimization with an initial learning
rate of 0.001.  For decoding, we use greedy search and beam search with a beam
size of 10.  As reported in \citet{bd_rnn_asr,Bahdanau2016}, we find that
penalizing candidate sentences that are too short was required to obtain the
best results. 
We add the middle term of Eq. \eqref{eq:augment_lm} to encourage longer candidate sentences.
All hyperparameters are chosen based on the development set. NPMT takes about 2--3
days to run to convergence (40 epochs) on a machine with four M40 GPUs. The results are summarized
in Table \ref{tab:iwslt_de-en}.  In addition to previous reported baselines in
the literature, we also explored the best hyperparameter using the same model
architecture (except the reordering layer) using sequence-to-sequence model with
attention as reported as LL$^{*}$ of Table \ref{tab:iwslt_de-en}. 

NPMT achieves state-of-the-art results on this dataset as far as we know.
Compared to the supervised sequence-to-sequence model, LL~\citep{Bahdanau2016},
NPMT achieves 2.4 BLEU gain in the greedy setting and 2.25 BLEU gain using
beam-search. Our results are also better than those from the actor-critic based
methods in~\citet{Bahdanau2016}. But we note that our proposed method is
orthogonal to the actor-critic method. So it is possible to further improve our
results using the actor-critic method. 

We also run the following two experiments to verify the sources of the gain.  The
first is to add a reordering layer to the original sequence-to-sequence model
with attention, which gives us BLEU scores of 25.55 (greedy) and 26.91 (beam
search). 
Since the attention mechanism and reordering layer capture similar information, adding the reordering layer to the sequence-to-sequence model with attention does not improve the performance.
The second is to remove the reordering layer from NPMT, which gives us
BLEU scores of 27.79 (greedy) and 29.28 (beam search). This shows that the
reordering layer and SWAN are both important for the effectiveness of NPMT.


%

\begin{table}[t!]
	\centering
	\begin{tabular}{lcc}
		\toprule		
		& \multicolumn{2}{c}{BLEU } \\ 
		&  Greedy & Beam Search \\ 	
		\midrule
		MIXER \citep{DBLP:journals/corr/RanzatoCAZ15} & 20.73 & 21.83 \\    
		LL \citep{wiseman2016sequence} & 22.53  & 23.87 \\
		BSO \citep{wiseman2016sequence} & {23.83} & {25.48} \\
		LL \citep{Bahdanau2016} & 25.82  & 27.56 \\
		LL$^{*}$ & 26.17  & 27.61 \\
		\midrule
		RF-C+LL \citep{Bahdanau2016} & {27.70} & {28.30} \\
		AC+LL \citep{Bahdanau2016} & {27.49} & {28.53} \\		
		\midrule
		
		NPMT (this paper) & \textbf{28.57} & \textbf{29.92} \\				
		NPMT+LM (this paper) & {--} & \textbf{30.08} \\				
		\bottomrule
	\end{tabular}
	\caption{\small{Translation results on the IWSLT 2014 German-English test set. MIXER
		\cite{DBLP:journals/corr/RanzatoCAZ15} uses a  convolutional encoder and
		simpler attention. LL (attention model with log likelihood) and BSO (beam
		search optimization) of \citet{wiseman2016sequence}, and LL, RF-C+LL, and
		AC+LL of \citet{Bahdanau2016} use a one-layer GRU encoder and decoder with
    attention. (RF-C+LL and AC+LL are different settings of actor-critic
    algorithms combined with LL.) LL$^{*}$ stands for a well-tuned attention
    model with log likelihood with the same word embedding size, and encoder and
    decoder size as NPMT.}
	}
	\label{tab:iwslt_de-en}
\end{table}

\begin{figure}[t!]
	\begin{center}
		\centerline{\includegraphics[width=0.75\columnwidth]{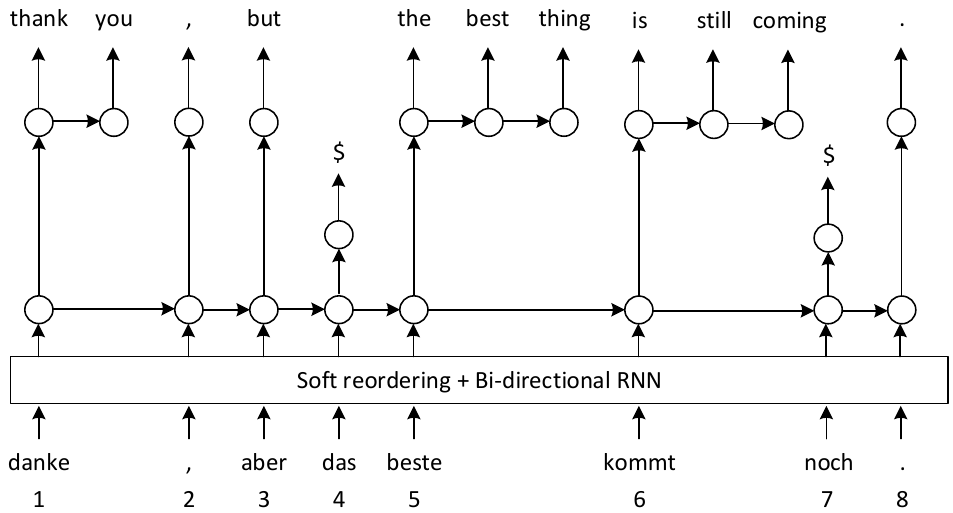}}
		\vskip -0.1in
		\caption{\small{An example of NPMT greedy decoding output for German-English
				translation. The example corresponds to the first example of Table
				\ref{tab:de-en_examples}. Note that for illustrating the input and output
				segments, we do not take into account of the behavior of the reordering
				layer and bi-directional RNN---the index mappings from source to target
				assumes monotonic alignments so some of them might be inaccurate.}}
		\label{fig:de-en_examples}
	\end{center}
	
\end{figure} 

\begin{table*}[h!] 
	\begin{small}
		{
			\begin{center}
				\begin{tabular}{rl}
					\hline

					
					source &  $^1$danke $^2$, $^3$aber $^4$das $^5$beste $^6$kommt $^7$noch $^8$.				
					\\
					greedy decoding& $^1$thank you $\bullet$ $^2$, $\bullet$ $^3$but $\bullet$ $^5$the best thing $\bullet$ $^6$is still coming $ \bullet$ $^8$.   \\
					target ground truth & thanks . i haven 't come to the best part .				
					\\			
					\hline
					source &  $^1$sie $^2$k\"onnen $^3$einen $^4$schalter $^5$dazwischen $^6$einf\"ugen $^7$und $^8$so $^9$haben \\& $^{10}$sie $^{11}$einen $^{12}$kleinen $^{13}$UNK $^{14}$erstellt $^{15}$.\\
					greedy decoding&  $^1$you can put $\bullet$ $^4$a switch $\bullet$ $^5$in between $\bullet$ $^7$and $\bullet$ $^8$so $\bullet$ $^{10}$they created $\bullet$ $^{12}$a little $\bullet$ $^{13}$UNK $^{14}$. \\
					target ground truth & you can put a knob in between and now you 've made a little UNK .
					\\			
					\hline
					source &  $^1$sie $^2$wollen $^3$die $^4$entscheidung $^5$wirklich $^6$richtig $^7$treffen $^8$, $^9$wenn $^{10}$es $^{11}$f\"ur \\& $^{12}$alle $^{13}$ewigkeit $^{14}$ist $^{15}$, $^{16}$richtig $^{17}$?
					\\
					greedy decoding&  $^1$you really want to make $\bullet$ $^4$the decision $\bullet$ $^6$right $\bullet$ $^8$, $\bullet$ $^9$if $\bullet$ $^{10}$it 's $\bullet$ $^{11}$for $\bullet$ \\& $^{12}$all $\bullet$ $^{13}$eternity $\bullet$ $^{15}$, $\bullet$ $^{16}$right $\bullet$ $^{17}$?   \\
					target ground truth & you really want to get the decision right if it 's for all eternity , right ?
					\\			
					\hline
					
					
					source & $^1$es $^2$gibt $^3$zehntausende $^4$maschinen $^5$rund $^6$um $^7$die $^8$welt $^9$die $^{10}$kleine $^{11}$st\"ucke $^{12}$von  \\ & $^{13}$dna $^{14}$herstellen $^{15}$k\"onnen $^{16}$,  $^{17}$30 $^{18}$bis $^{19}$50  $^{20}$buchstaben  $^{21}$lang $^{22}$aber  $^{23}$es \\ &$^{24}$ist $^{25}$ein $^{26}$UNK $^{27}$prozess $^{28}$, $^{29}$also $^{30}$je $^{31}$l\"anger $^{32}$man $^{33}$ein $^{34}$st\"uck $^{35}$macht $^{36}$,\\ & $^{37}$umso $^{38}$mehr $^{39}$fehler $^{40}$passieren $^{41}$.							
					\\
					greedy decoding&   $^1$there are $\bullet$  $^3$tens of thousands of $\bullet$  $^4$machines $\bullet$  $^6$around $\bullet$  $^8$the world $\bullet$  $^9$can make $\bullet$ \\ & $^{10}$little $\bullet$ $^{11}$pieces $\bullet$ $^{12}$of $\bullet$ $^{13}$dna $\bullet$ $^{16}$,  $\bullet$  $^{17}$30 $\bullet$ $^{18}$to $\bullet$ $^{19}$50 $\bullet$ $^{20}$letters $\bullet$ $^{21}$long $\bullet$ $^{22}$,  \\ &but $\bullet$ $^{23}$it 's $\bullet$ $^{26}$a more UNK $\bullet$ $^{27}$process $\bullet$ $^{28}$, $\bullet$ $^{29}$so $\bullet$ $^{31}$the longer $\bullet$ $^{32}$you make $\bullet$ \\ &$^{34}$a piece $\bullet$ $^{36}$,  $\bullet$ $^{38}$the more $\bullet$ $^{39}$mistakes $\bullet$ $^{40}$happen $\bullet$ $^{41}$.    \\
					target ground truth & there are tens of thousands of machines around the world that make small pieces of dna \\ &-- 30 to 50 letters - in length - and it 's a UNK process , \\ &so the longer you make the piece , the more errors there are .								
					\\			
					\hline
				\end{tabular}
			\end{center}
		}
	\end{small}
	\vspace{-2mm}
	\caption{\small{Examples of German-English translation outputs with their
			segmentations. We label the indexes of the words in the source sentence and we
			use those indexes to indicate where the output segment is emitted. For
			example, in greedy decoding results, ``$^i \texttt{word}_1, \ldots,
			\texttt{word}_m$'' denotes $i$-th word in the source sentence emits words
			$\texttt{word}_1, \ldots, \texttt{word}_m$ during decoding (assuming monotonic
			alignments).  The ``$\bullet$'' represents the segment boundary in the target
			output. See Figure \ref{fig:de-en_examples} for a visualization of row 1 in this table.}}
	\label{tab:de-en_examples}
\end{table*}

In greedy decoding, we can estimate the {\it average segment
length}\footnote{The average segment length is defined as the length of the
output (excluding end of segment symbol $\$$) divided by the number of segments
(not counting the ones only containing $\$$).} for the output.  The average
segment length is around 1.4--1.6, indicating phrases with more than one word
are being decoded.  Figure \ref{fig:de-en_examples} shows an example of the input and
decoding results with NPMT.  We can observe phrase-level translation being
captured by the model (e.g., ``danke'' $\rightarrow$ ``thank you''). The model
also knows when to {\it sleep} before outputting a phrase (e.g., ``das''
$\rightarrow$ ``\$'').  We use the indexes of words in the source sentence to indicate
where the output phrases are from.  Table \ref{tab:de-en_examples} shows some
sampled examples.  We can observe there are many informative segments
in the decoding results, e.g., ``tens of thousands of'', ``the best thing'', ``a
little'', etc.  There are also mappings from phrase to phrase, word to phrases,
and phrase to word in the examples.  Following the analysis, we show the most
frequent phrase mappings in Appendix \ref{sec:phrase_mapping}.
 
We also explore an option of adding a language-model score during beam search as
the traditional statistical machine translation does.  This option might not
make much sense in attention-based approaches, since the decoder itself is
usually a neural network language model. In SWAN, however, there is no language
models directly involved in the segmentation
modeling,\footnote{In~\citet{wang2017sequence}, SWAN does have an option to use a
separate RNN that connects the segments, which can be seen as a language model.
However, different from speech recognition experiments, we find in machine
translation experiments, adding this separate RNN leads to a worse performance.
We suspect this is because an RNN language model can be easier to learn than
the segmentation structures and SWAN gets stuck in that local mode.  This is
further evidenced by the fact that the average segment length is much shorter
with a separate RNN in SWAN.} and we find it useful to have an external language
model during beam search. We use a 4th-order language model trained using the KenLM
implementation~\citep{Heafield-estimate} for English target training data. So
the final beam search score we use is 
\begin{equation}
 Q(y) = \log p(y|x)  + \lambda_1 {\rm
  word\_count}(y)+ \lambda_2 \log p_{\rm lm}(y), 
\label{eq:augment_lm}
\end{equation}
where we empirically find that $\lambda_1=1.2$ and $\lambda_2=0.2$ give good
performance, which are tuned on the development set. The results with the external language model are denoted by NPMT+LM in Table \ref{tab:iwslt_de-en}. If no external language
models are used, we set $\lambda_2=0$. This scoring function is similar to the
one for speech recognition in~\citet{hannun2014deep}.

\subsection{IWSLT14 English-German}
\label{sec:iwslt_en-de}
We also evaluate our model on the opposition direction, English-German, which
translates from a more segmented text to a more inflectional one.  Following the
setup in Section \ref{sec:iwslt_de-en}, we use the same dataset with the
opposite source and target languages.  We use the same model architecture,
optimization algorithm and beam search size as the German-English translation
task. NPMT takes about 2--3 days to run to convergence (40 epochs) on a machine with four M40 GPUs.

Given there is no previous sequence-to-sequence attention model baseline for
this setup, we create a strong one and tune hyperparameters on the development
set.  The results are shown in Table \ref{tab:iwslt_en-de}. Based on the
development set, we set $\lambda_1=1$ and $\lambda_2=0.15$ in Eq.
\eqref{eq:augment_lm}.  Our model outperforms sequence-to-sequence model with
attention by 2.46 BLEU and 2.49 BLEU in greedy and beam search cases.  We can
also use a 4th-order language model trained using the KenLM
implementation for German target training data, which
further improves the performance.  Some sampled examples are shown in
Table \ref{tab:en-de_examples}.  Several informative segments/phrases can be
found in the decoding results, e.g., ``some time ago'' $\rightarrow$ ``vor
enniger zeit''.

\begin{table}[h]
	\centering
	\begin{tabular}{lcc}
		\toprule		
		& \multicolumn{2}{c}{BLEU } \\ 
		&  Greedy & Beam Search \\ 	
		\midrule
    Sequence-to-sequence with attention \cite{} & 21.26  & 22.59 \\
		NPMT (this paper) & \textbf{23.62} & \textbf{25.08} \\	
	    NPMT+LM (this paper) & {--} & \textbf{25.36} \\									
		\bottomrule
	\end{tabular}
	\caption{\small{Translation results on the IWSLT 2014 English-German
		test set.}}
	\label{tab:iwslt_en-de}
\end{table}

\begin{table*}[h!] 
	
	\begin{small}
		{
			\begin{center}
				\begin{tabular}{rl}
					\hline
					source &  $^1$how $^2$would $^3$you $^4$guys $^5$describe $^6$your $^7$brand $^8$?
					\\
					greedy decoding& $^1$wie $\bullet$ $^2$w\"urdet $\bullet$ $^3$sie $\bullet$ $^6$ihre marke $\bullet$ $^8$beschreiben ? \\
					target ground truth & wie w\"urdet ihr eure marke beschreiben ?				
					\\			
					\hline
					source &  $^1$if $^2$the $^3$museum $^4$has $^5$given $^6$us $^7$the $^8$image $^9$, $^{10}$you $^{11}$click $^{12}$on $^{13}$it $^{14}$.\\
					greedy decoding&  $^1$wenn $\bullet$ $^2$das museum $\bullet$ $^6$uns $\bullet$ $^7$das bild $\bullet$ $^9$gegeben hat ,$\bullet$ $^{10}$klicken sie $\bullet$ $^{13}$darauf $\bullet$ $^{14}$. \\
					target ground truth & wenn das museum uns das bild gegeben hat , klicken sie darauf .
					\\			
					\hline
					source &  $^1$they $^2$are $^3$frustrated $^4$as $^5$hell $^6$with $^7$it $^8$, $^9$but $^{10}$they $^{11}$'re  $^{12}$not $^{13}$complaining\\& $^{14}$about $^{15}$it $^{16}$, $^{17}$they $^{18}$'re $^{19}$fixing $^{20}$it $^{21}$.
					\\
					greedy decoding&  $^1$sie sind $\bullet$ $^3$frustriert $\bullet$ $^8$, $\bullet$ $^9$aber $\bullet$ $^{10}$sie UNK sich $\bullet$ $^{12}$nicht $\bullet$ \\& $^{15}$dar\"uber $\bullet$ $^{16}$,  $\bullet$ $^{17}$sie reparieren $\bullet$ $^{20}$es $\bullet$ $^{21}$.   \\
					target ground truth & sie sie sind f\"urchterlich frustriert mit ihr , aber sie beschweren sich nicht dar\"uber ,\\& sie reparieren sie . ?\\			
					\hline			
					source & $^1$now $^2$some $^3$time $^4$ago $^5$, $^6$if $^7$you $^8$wanted $^9$to $^{10}$win $^{11}$a $^{12}$formula  \\ & $^{13}$1 $^{14}$race $^{15}$, $^{16}$you  $^{17}$take $^{18}$a $^{19}$budget  $^{20}$,  $^{21}$and $^{22}$you  $^{23}$bet \\ &$^{24}$your $^{25}$budget $^{26}$on $^{27}$a $^{28}$good $^{29}$driver $^{30}$and $^{31}$a $^{32}$good $^{33}$car $^{34}$.					
					\\
					greedy decoding&   $^2$vor einiger zeit $\bullet$  $^6$wenn $\bullet$  $^7$man $\bullet$  $^{11}$eine formel $\bullet$  $^{15}$gewinnen will , $\bullet$  $^{18}$ein budget $\bullet$ \\ & $
					
					^{21}$und $\bullet$ $^{23}$ , dass $\bullet$ $^{24}$ihr budget $\bullet$ $^{27}$auf einem guten $\bullet$ $^{29}$fahrer  $\bullet$  $^{30}$und $\bullet$ $^{31}$ein gutes $\bullet$ \\& $^{33}$auto $\bullet$ $^{34}$.\\
					target ground truth & vor einiger zeit war es so , dass wenn sie ein formel 1 rennen gewinnen wollten , \\& dann nahmen sie ihr budget und setzten ihr geld auf einen guten fahrer und ein gutes auto . 					
					\\			
					\hline
				\end{tabular}
			\end{center}
		}
	\end{small}
  \caption{\small{Examples of English-German translation outputs with their
  segmentations.  The meanings of the superscript indexes and the ``$\bullet$''
  symbol are the same as those in Table~\ref{tab:de-en_examples}.}}
  \label{tab:en-de_examples}
\end{table*}

\subsection{IWSLT15 English-Vietnamese}
In this section, we evaluate our model on the IWSLT 2015 English to Vietnamese
machine translation task.
The data is from translated TED talks, and the dataset contains roughly 133K
training sentence pairs provided by the IWSLT 2015 Evaluation
Campaign~\citep{cettolo2015iwslt}.  Following the same preprocessing steps in
\cite{luong_mt_2015,raffel2017online}, we use the TED tst2012 (1553 sentences)
as a validation set for hyperparameter tuning and TED tst2013 (1268 sentences)
as a test set.
The Vietnamese and English vocabulary sizes are 7,709 and 17,191 respectively.

We use one reordering layer with window size 7, two layers of bi-directional LSTM (Long
short-term memory, \cite{Hochreiter:1997}) encoder with 512 hidden units, and
three layers of unidirectional LSTM decoder with 512 hidden units.  We add dropout with a
rate of $0.4$ in the LSTM layer. We choose LSTM since baselines for comparisons
were using LSTM. The maximum segment length is set to 7.  Batch size is set as 48 (per GPU)
and the Adam algorithm~\cite{kingma2014adam} is used for optimization with an
initial learning rate of 0.001. For decoding, we use greedy decoding and beam
search with a beam size of 10. The results are shown in Table
\ref{tab:iwslt_en-vi}.  Based on the development set, we set $\lambda_1=0.7$ and
$\lambda_2=0.15$ in Eq. \eqref{eq:augment_lm}.  
NPMT takes about one day to run to convergence (15 epochs) on a machine with 4 M40 GPUs. 
Our model outperforms sequence-to-sequence model with attention by 1.41 BLEU and 1.59 BLEU in greedy
and beam search cases.  We also use a 4th-order language model trained using the KenLM
implementation for Vietnamese target training data,
which further improves the BLEU score. Note that our reordering layer relaxes
the monotonic assumption as in \cite{raffel2017online} and is able to decode in
linear time. Empirically we outperform models with monotonic attention.  Table
\ref{tab:en_vi_examples} shows some sampled examples. 
\begin{table}[h!]
	\centering
	\begin{tabular}{lcc}
		\toprule		
		& \multicolumn{2}{c}{BLEU } \\ 
		&  Greedy & Beam Search \\ 	
		\midrule

		Hard monotonic \citep{raffel2017online} & 23.00  & - \\
		\citet{luong_mt_2015} & - &  23.30 \\    
		Sequence-to-sequence model with attention & 25.50 &  26.10 \\  
		NPMT (this paper) & \textbf{26.91} & \textbf{27.69} \\				
		NPMT+LM (this paper) & -- & \textbf{28.07} \\				
		\bottomrule
	\end{tabular}
  \caption{\small{Translation results on the IWSLT 2015 English-Vietnamese tst2013 test
  set. The result of the sequence-to-sequence model with attention is obtained
  from an open source model provided by the authors.\protect\footnotemark}}
	\label{tab:iwslt_en-vi}
\end{table}

\begin{table*}[th!] 
	\begin{small}
		{
			\begin{center}
				\begin{tabular}{rl}
					\hline
					source &  $^1$And $^2$I $^3$figured $^4$, $^5$this $^6$has $^7$to $^8$stop $^9$.
					\\
					greedy decoding & $^1$V\`a $\bullet$ $^2$t\^oi $\bullet$ $^3$nh\d\acircumflex n ra r\`\abreve ng $\bullet$ $^4$, $\bullet$ $^5$\dj i\`\ecircumflex u n\`ay $\bullet$ $^6$ph\h ai$\bullet$ $^8$d\`\uhorn  ng  l\d ai $^9$. \\
					target ground truth & V\`a t\^oi nh\d\acircumflex n ra r\`\abreve ng  \dj i\`\ecircumflex u \dj\'o ph\h ai ch\'\acircumflex m d\h\uhorn t . 
					\\	\hline
					source &  $^1$So $^2$great $^3$progress $^4$and $^5$treatment $^6$has $^7$been $^8$made $^9$over $^{10}$the $^{11}$years $^{12}$.
					\\
					greedy decoding & $^1$V\`i v\d \acircumflex y , $\bullet$ $^2$ti\'\ecircumflex n b\d \ocircumflex $\bullet$ $^4$v\`a $\bullet$ $^5$\dj i\`\ecircumflex u tr\d i $\bullet$ $^6$\dj \~a $\bullet$ $^7$\dj \uhorn \d \ohorn c  $\bullet$ $^8$t\d ao ra  $\bullet$ $^9$trong  $\bullet$ $^{10}$nh\~\uhorn ng  $\bullet$ \\& $^{11}$n\abreve m $\bullet$ $^{12}$.  \\
					target ground truth & Trong su\'\ocircumflex t nh\~\uhorn ng n\abreve m qua \dj \~a c\'o s\d \uhorn~ ti\'\ecircumflex n b\d \ocircumflex ~to l\'\ohorn n trong qu\'a tr\`inh \dj i\`\ecircumflex u tr\d i .
					\\	\hline

	source &  $^1$The $^2$passion $^3$that $^4$the $^5$person $^6$has $^7$for $^8$her $^9$own $^{10}$growth $^{11}$is $^{12}$the \\& $^{13}$most $^{14}$ important $^{15}$ thing $^{16}$.
\\
greedy decoding & $^1$Ni\`\ecircumflex m \dj am m\^e $\bullet$ $^3$r\`\abreve ng $\bullet$ $^5$ng\uhorn \`\ohorn
i $\bullet$ $^6$ c\'o $\bullet$ $^7$ cho $\bullet$ 
$^{8}$s\d \uhorn ~ph\'at tri\h \ecircumflex n $\bullet$ $^{10}$c\h ua c\ocircumflex ~\'\acircumflex y $\bullet$ $^{11}$l\`a $\bullet$ \\& $^{13}$\dj i\`\ecircumflex u $\bullet$ $^{14}$quan tr\d ong $\bullet$  $^{15}$nh\'\acircumflex t $\bullet$ $^{16}$.
\\
target ground truth & 	C\'ai kh\'at v\d ong c\h ua ng\uhorn\`\ohorn i ph\d u n\~\uhorn~c\'o cho s\d \uhorn ~ph\'at tri\h \ecircumflex n c\h ua b\h an th\acircumflex n l\`a th\'\uhorn ~quan tr\d ong nh\'\acircumflex t .
\\	\hline
					

				source &  $^1$We $^2$have $^3$eight $^4$species $^5$of $^6$UNK $^7$that $^8$occur $^9$ in $^{10}$Kenya $^{11}$, $^{12}$of $^{13}$which $^{14}$six \\& $^{15}$are $^{16}$highly $^{17}$threatened $^{18}$with $^{19}$ extinction $^{20}$.
				\\
				greedy decoding & $^1$Ch\'ung ta $\bullet$ $^2$c\'o $\bullet$ $^3$8 $\bullet$ $^4$lo\`ai $\bullet$ $^6$ UNK $\bullet$ $^{8}$x\h ay ra $\bullet$ $^{9}$\h\ohorn~$\bullet$ $^{10}$Kenya $\bullet$ $^{11}$,~$\bullet$ $^{14}$6 $\bullet$ \\ & $^{17}$b\d i  \dj e do\d a $\bullet$ 
				$^{19}$tuy\d \ecircumflex t ch\h ung $\bullet$ $^{20}$.\\
				target ground truth &  Ch\'ung ta c\'o 8 lo\`ai k\`\ecircumflex n k\`\ecircumflex n xu\'\acircumflex t hi\d \ecircumflex n t\d ai Kenya , trong \dj\'o c\'o 6 lo\`ai b\d i \dj e do\d a v\'\ohorn i nguy \\ & c\ohorn~ tuy\d \ecircumflex t ch\h ung cao .
				\\

					\hline
				\end{tabular}
			\end{center}
		}
	\end{small}
  \caption{\small{Examples of English-Vietnamese translation outputs with their
  segmentations.  The meanings of the superscript indexes and the ``$\bullet$''
  symbol are the same as those in Table~\ref{tab:de-en_examples}.}}
  \label{tab:en_vi_examples}
\end{table*}
\footnotetext{https://github.com/tensorflow/nmt}
%

\section{Conclusion}
\label{sec:conclusion}
We proposed NPMT, a neural phrase-based machine translation system that models
phrase structures in the target language using SWAN. We also introduced a local
reordering layer to mitigate the monotonic alignment requirement in SWAN.  Our
experimental results showed promising results on IWSLT 2014 German-English,
English-German, and IWSLT 2015 English-Vietnamese machine translation tasks. 
The results suggest that NPMT can potentially be extended to explore the structures in other challenging sequence-to-sequence problems.
In future work, we will explore two directions: 1) speed up NPMT and apply it to larger datasets
and more language pairs; 2) investigate how to learn input and output phrases
simultaneously.

\section{Acknowledgments}
We thank Jacob Devlin, Adith Swaminathan, Frank Seide, Xiaodong He, and anonymous reviewers for their valuable feedback. 

\bibliography{bib}

\bibliographystyle{iclr2018_conference}

\clearpage
\appendix

\section{Reordering Layer Analysis}
\label{sec:reordering_analysis}
\begin{figure}[h]
	\begin{center}
		\centerline{\includegraphics[width=0.98\columnwidth]{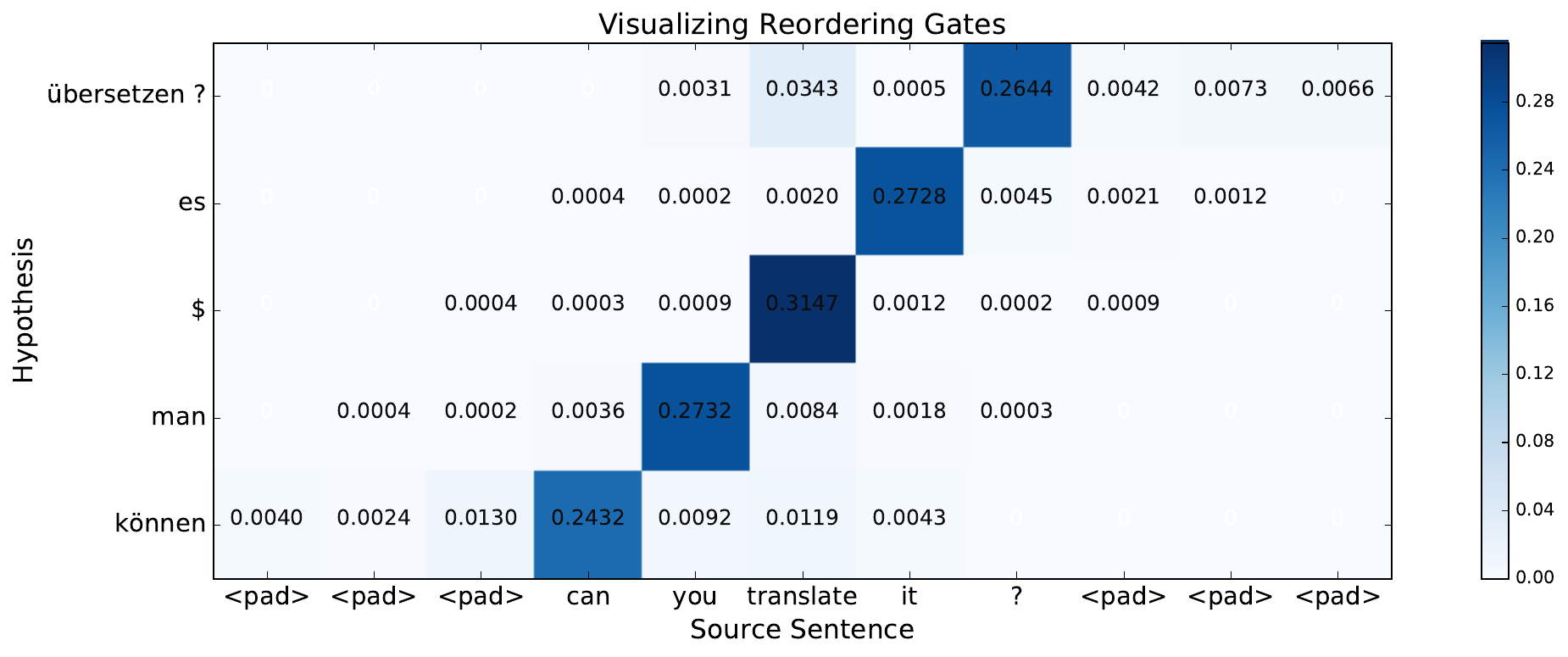}}
		\caption{\small{Visualizing reordering gates in the NPMT English-German translation model.}}
		\label{fig:reordering_analysis_en-de}
	\end{center}
\end{figure}

To further understand the behavior of the reordering layer, we examine the
values of the gate $\sigma\left(w_i^T [e_{t-\tau};\ldots;e_t; \ldots;
e_{t+\tau}]\right)$ in Eq. \eqref{eq:reordering}.  We study the NPMT
English-German model in Section \ref{sec:iwslt_en-de}.  In Figure
\ref{fig:reordering_analysis_en-de}, we show an example that translates from
``can you translate it ?'' to ''k\"onnen man es \"ubersetzen ?'', where the
mapping between words are as follows: ``can $\rightarrow$ k\"onnen'', ``you
$\rightarrow$ man'', ``translate $\rightarrow$ \"ubersetzen'', ``it
$\rightarrow$ es'' and ``? $\rightarrow$ ?''.  Note that the example needs to be
reordered from ``translate it'' to ''es \"ubersetzen''.  Each row of Figure
\ref{fig:reordering_analysis_en-de} represents a window of size 7 that is
centered at a source sentence word. The values in the matrix represent	the gate
values for the corresponding words.  The gate values will later be multiplied
with the embedding $e_{t-\tau+i}$ of  Eq. \eqref{eq:reordering} and contribute
to the hidden vector $h_t$.  The y-axis represents the word/phrases emitted from
the corresponding position.  We can observe that the gates mostly focus on the
central word since the first part of the sentence only requires monotonic
alignment. Interestingly, the model outputs ``\$'' (empty) when the model has
the word ``translate'' in the center of the window.  Then, the model outputs
``es'' when the model encounters ``it''.  Finally, in the last window (top row),
the model not only has a large gate value to the center input ``?'', but the
model also has a relatively large gate value to the word ``translate'' in order
to output the translation ``\"ubersetzen ?''.  This shows an example of the
reordering effect achieved by using the gating mechanism of the reordering
layer.

\section{Effect of Window Sizes in the Reordering Layer}
\label{sec:window_reorder}
In this section, we examine the effect of window sizes in the reordering layer.
Following the setup in Section \ref{sec:iwslt_en-de}, we evaluate the
performance of different window sizes on the IWSLT 2014 English-German
translation task. Table \ref{tab:iwslt_en-de_window_sizes} summarizes the
results. We can observe that the performance reaches the peak with a windows size of
7.  With a window size of 5, the performance drops 0.88 BLEU in greedy decoding and
0.72 BLEU using beam search. It suggests that the context window is not large
enough to properly perform reordering.  When the window sizes are 9 and 11, we
do not observe further improvements.  It might be because the translation between English and German mostly requires local word reordering.

\begin{table}[h]
	\centering
	\begin{tabular}{ccc}
		\toprule		
		& \multicolumn{2}{c}{BLEU } \\ 
		Window Size &  Greedy & Beam Search \\ 	
		\midrule
		5 & 22.74 & 24.36 \\
		7 & {\bf 23.62} & {\bf 25.08} \\
		9 & 23.11 & 24.68 \\
		11 & 23.12 & 24.65 \\
		\bottomrule
	\end{tabular}
	\caption{\small{Analyze the effect of reordering layer window sizes in translation results on the IWSLT 2014 English-German test set.}}
	\label{tab:iwslt_en-de_window_sizes}
\end{table}

\section{Phrase mapping examples}
\label{sec:phrase_mapping}
Following the examples of Table \ref{tab:de-en_examples}, we analyze the
decoding results on the test set of the German-English translation task.  Given
we do not have explicit input segments in NPMT, we assume input words that emit
``\$'' symbol are within the same group as the next non-'\$' word.  For example,
in Figure \ref{fig:de-en_examples}, input words ``das beste'' are considered as
an input segment.  We then can aggregate all the input, output segments
(phrases) and sort them based on the frequency.  Tables
\ref{tab:de-en_phrase_table} and \ref{tab:de-en_phrase_table_long} show the
most-frequent input, output phrase mappings.

\begin{table}[h]
  \vskip 0.1in
\label{tab:de-en_phrase_table}
    \hspace{-6mm}
	\begin{tabular}{|l|l|l|l|l|}
		\hline
		\multicolumn{1}{|c|}{\textbf{One $\rightarrow$ One}} & \multicolumn{1}{c|}{\textbf{One $\rightarrow$ Many}} & \multicolumn{1}{c|}{\textbf{Many $\rightarrow$ One}} & \multicolumn{1}{c|}{\textbf{Many $\rightarrow$ Many}} & \multicolumn{1}{c|}{\textbf{Many $\rightarrow$ Many$^*$}} \\ \hline
		, $\rightarrow$ ,                                    & es $\rightarrow$ it 's                               & , dass $\rightarrow$ that                            & die UNK $\rightarrow$ the UNK                         & wissen sie $\rightarrow$ you know                              \\ \hline
		. $\rightarrow$ .                                    & UNK $\rightarrow$ the UNK                            & in der $\rightarrow$ in                              & der UNK $\rightarrow$ the UNK                         & in diesem $\rightarrow$ in this                                \\ \hline
		und $\rightarrow$ and                                & und $\rightarrow$ , and                              & UNK . $\rightarrow$ .                                & ein UNK $\rightarrow$ a UNK                           & die welt $\rightarrow$ the world                               \\ \hline
		UNK $\rightarrow$ UNK                                & das $\rightarrow$ this is                            & UNK , $\rightarrow$ ,                                & das UNK $\rightarrow$ the UNK                         & ist es $\rightarrow$ it 's                                     \\ \hline
		aber $\rightarrow$ but                               & das, $\rightarrow$ that 's                           & , die $\rightarrow$ that                             & eine UNK $\rightarrow$ a UNK                          & '' . $\rightarrow$ . ''                                             \\ \hline
		'' $\rightarrow$ ''                                       & UNK $\rightarrow$ a UNK                              & ist . $\rightarrow$ .                                & in UNK $\rightarrow$ in UNK                           & ein paar $\rightarrow$ a few                                   \\ \hline
		ist $\rightarrow$ is                                 & ich $\rightarrow$ i think                            & in den $\rightarrow$ in                              & den UNK $\rightarrow$ the UNK                         & gibt es $\rightarrow$ there 's                                 \\ \hline
		der $\rightarrow$ of                                 & es $\rightarrow$ it was                              & ist , $\rightarrow$ ,                                & wissen sie $\rightarrow$ you know                     & der welt $\rightarrow$ the world                               \\ \hline
		von $\rightarrow$ of                                 & dies $\rightarrow$ this is                           & sind . $\rightarrow$ .                               & in diesem $\rightarrow$ in this                       & die frage $\rightarrow$ the question                           \\ \hline
		mit $\rightarrow$ with                               & es $\rightarrow$ there 's                            & , wenn $\rightarrow$ if                              & dem UNK $\rightarrow$ the UNK                         & haben wir $\rightarrow$ we have                                \\ \hline
	\end{tabular}
\caption{\small{German-English phrase mapping results. We show the top 10 input,
		output phrase mappings in five categories (``One'' stands for single word and
		``Many'' stands for multiple words.). In the last column, Many $\rightarrow$
		Many$^*$, we remove the phrases with the ``UNK'' word as the ``UNK'' appears
		often.}}
\end{table}

\begin{table}[h]
	\centering

  \vskip 0.1in
	\label{tab:de-en_phrase_table_long}
\begin{tabular}{|l|l|}
	\hline
	\multicolumn{1}{|c|}{\textbf{Phrases with 3 words}} & \multicolumn{1}{c|}{\textbf{Phrases with 4 words}} \\ \hline
	auf der ganzen $\rightarrow$ all over the             & auf der ganzen $\rightarrow$ a little bit of         \\ \hline
	gibt eine menge $\rightarrow$ a lot of                & wei\ss~nicht , was $\rightarrow$ what 's going to be \\ \hline
	dann hat er$\rightarrow$ he doesn 't have             & tun , das wir $\rightarrow$  we can 't do            \\ \hline
	, die man $\rightarrow$ you can do                    & tat , das ich $\rightarrow$ i didn 't do             \\ \hline
	das k\"onnen wir $\rightarrow$ we can do that         & zu verbessern , die $\rightarrow$ that can be done   \\ \hline
	\end{tabular}
  \caption{\small{German-English longer phrase mapping results. We show the top 5
	input, output phrase mappings for two categories: input and output phrases
	with three words, and input and output phrases with four words.}}
\end{table}

\end{document}